\documentclass[graybox]{svmult}

\usepackage{type1cm}        
\usepackage{makeidx}         
\usepackage{graphicx}        
\usepackage{multicol}        
\usepackage[bottom]{footmisc}

\usepackage{newtxtext}       %
\usepackage{newtxmath}       

\usepackage{algorithm2e}

\DeclareFontFamily{OMS}{ntxtt}{\hyphenchar\font45 }
\DeclareFontFamily{OMS}{ntxtlf}{\hyphenchar\font45 }
\DeclareFontShape{OMS}{ntxtt}{m}{n}{<->ssub*cmsy/m/n}{}
\DeclareFontShape{OMS}{ntxtlf}{m}{n}{<->ssub*cmsy/m/n}{}

\usepackage[style=apa, backend=biber]{biblatex}
\usepackage{cellspace}

\usepackage{alphabeta}

\makeindex             

\begin{document}

\title*{Conceptual engineering using large language models}

\author{Bradley P. Allen}

\institute{Bradley P. Allen \at University of Amsterdam, Amsterdam, The Netherlands, \email{b.p.allen@uva.nl}}
 
\maketitle

\abstract{We describe a method, based on Jennifer Nado's proposal for classification procedures as targets of conceptual engineering, that implements such procedures by prompting a large language model. We apply this method, using data from the Wikidata knowledge graph, to evaluate stipulative definitions related to two paradigmatic conceptual engineering projects: the International Astronomical Union's redefinition of PLANET and Haslanger's ameliorative analysis of WOMAN. Our results show that classification procedures built using our approach can exhibit good classification performance and, through the generation of rationales for their classifications, can contribute to the identification of issues in either the definitions or the data against which they are being evaluated.  We consider objections to this method, and discuss implications of this work for three aspects of theory and practice of conceptual engineering: the definition of its targets, empirical methods for their investigation, and their practical roles. The data and code used for our experiments, together with the experimental results, are available in a Github repository\footnote{\url{https://github.com/bradleypallen/zero-shot-classifiers-for-conceptual-engineering}}.}

\section{Introduction}
\label{sec:1}

\textit{Conceptual engineering} is a philosophical methodology concerned with "the design, implementation, and evaluation of concepts" \parencite{chalmers2020conceptual}. The goals of conceptual engineering are varied, e.g., achieving greater clarity and precision in argumentation and scientific discourse \parencite{justus2012carnap,dutilh2017carnapian}, or altering terminology to advance the cause of social justice \parencite{haslanger2000gender,manne2017down,podosky2022can}. Conceptual engineering projects often begin with an examination of the meaning and connotations of one or more natural language terms denoting a specific concept, addressing how those terms are used in the context of communicative exchanges between speakers of a given language \parencite{etta2021conceptual} and identifying how and why the concept is in need of revision. Proposals for new concepts, or for changes to an existing concept, are expressed and argued for in natural language. One major criterion for the success of a conceptual engineering project is if it leads to speakers using terms in a manner that reflects the engineered concept \parencite{pinder2022haslanger}. The methodological debates about the proper conduct of conceptual engineering are conducted through linguistic analysis and argumentation \parencite{burgess2020conceptual}. Philosophers have proposed differing theories as to how conceptual engineering is best defined and practiced, but it is clearly an activity where the use and analysis of natural language plays a significant role.

In recent years, large language models (LLMs) have emerged as a technology that promises to be of "substantial value in the scientific study of language learning and processing" \parencite{mahowald2023dissociating}. Given this, we ask the question: might LLMs be useful in the conduct of conceptual engineering projects? In this paper, we argue that that is the case.

The structure of this paper is as follows: we begin by describing different theories about the targets of conceptual engineering, focusing on a specific theory of Jennifer Nado. We then show how prompt programming of an LLM can be used to implement a classification procedure. We then show a way to evaluate such classification procedures using data from a knowledge graph, and conduct experiments based on two paradigmatic examples of conceptual engineering projects. We then discuss the results of the experiments from several perspectives: objections that could be raised to the use of LLMs in this  manner, and ways in which our method could address several issues in the theory and practice of conceptual engineering.

\section{Classification procedures as targets of conceptual engineering}
\label{sec:2}
\textcite{koch2023recent} surveys recent work on the theory of conceptual engineering, and identifies two core components of any such theory:
\begin{itemize}
    \item A theory of \textit{targets}: \textit{what} conceptual engineering creates or changes.
    \item A theory of \textit{engineering}: \textit{how} conceptual engineering is performed.
\end{itemize}

Much of the discussion in recent years around the theory of conceptual engineering has centered on responses to Herman Cappelen's Austerity Framework \parencite{cappelen2018fixing}, in which he defines conceptual engineering as "the practice of trying to change the extensions of linguistic items via changes in their intension" \parencite{jorem2024inferentialist}, i.e., that the targets of conceptual engineering are intensions. Alternative proposals for the targets of conceptual engineering range from the meanings speakers assign to terms \parencite{pinder2021conceptual}, psychological structures such as prototypes \parencite{isaac2022conceptual}, pluralistic approaches integrating both semantic meanings and psychological concepts \parencite{koch2021engineering}, and social norms such as entitlements \parencite{thomasson2020pragmatic,kohler2024conceptual}.\footnote{It is beyond the scope of this paper to provide a thorough discussion of these alternatives; for that, see \textcite{koch2023recent} and \textcite{burgess2020conceptual}.} \textcite{belleri2021pluralism} suggests that, given this range of proposals, a pluralist stance towards the targets of conceptual engineering is appropriate.

In this work we focus on the proposal for targets of conceptual engineering in \textcite{nado2021classification}: 
\begin{quote}
    A \textit{classification procedure} is any procedure that, when followed, allows the user to sort a set of entities into two groups--- those 'in' the category delineated by the procedure, and those 'out' of that category. 'Procedure' here is used in the ordinary English sense; a procedure is a method, a process, a set of steps aimed at achieving a goal \parencite[12]{nado2021classification}.
\end{quote}
From the perspective of a practitioner of artificial intelligence or machine learning, this is a very general way of describing a binary classifier. There are a plethora of ways in which one can create binary classifiers, but in the context of conceptual engineering, concepts are usually described using stipulative definitions in natural language. The natural language processing capabilities of LLMs and their successful application to text classification tasks \parencite{fields2024survey} suggests the possibility of implementing classification procedures as computational artifacts in a manner consistent with this practice, i.e., that conceptual engineers could create a classification procedure simply by providing an intensional definition of a concept in natural language.

\section{Constructing classification procedures using LLMs}
\label{sec:3}
\begin{figure*}
    \includegraphics[width=\columnwidth]{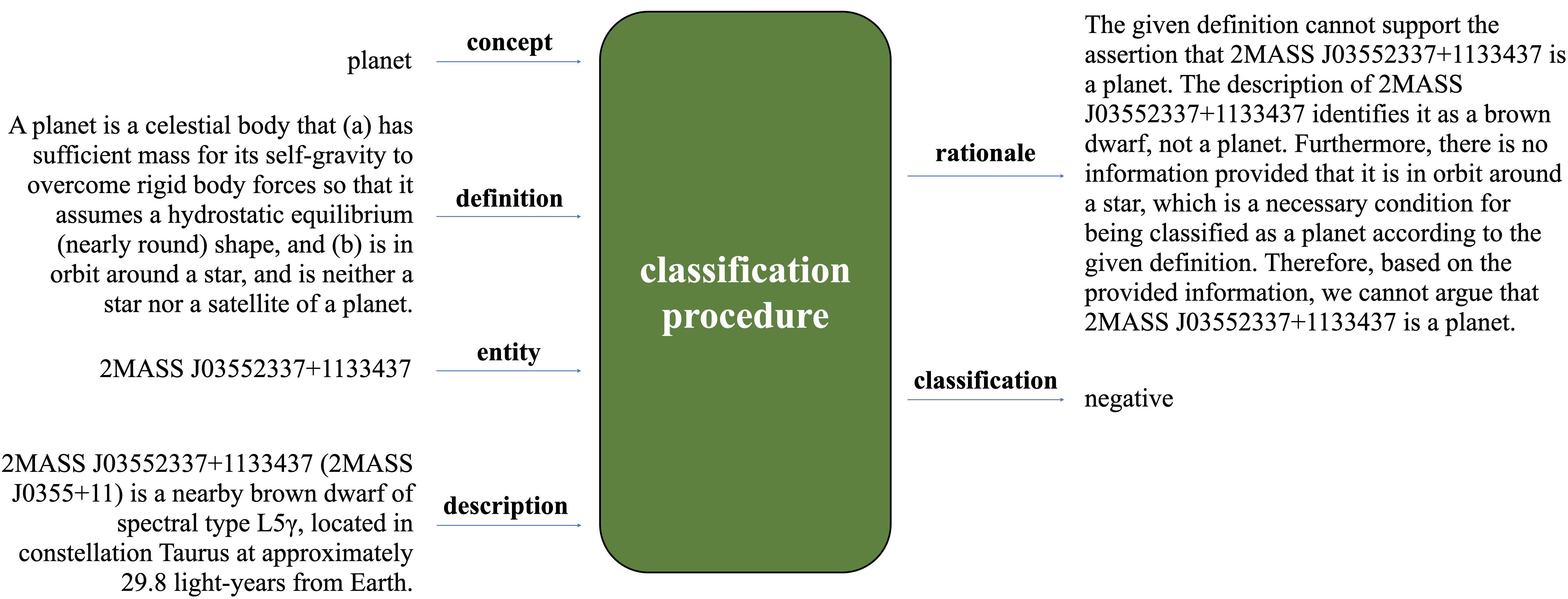}
    \caption{A classification procedure using the 24 August 2006 version of the IAU definition of PLANET, implemented as a zero-shot chain-of-thought classifier, and being applied to the description of the entity 2MASS J03552337+1133437.}
    \label{fig:classifier_example}
\end{figure*}

To accomplish this, we define a classification procedure as a zero-shot chain-of-thought classifier \parencite{kojima2022large}. Figure \ref{fig:classifier_example} shows an example of such a classification procedure. Given a concept's name and intensional definition and an entity's name and description, we prompt an LLM to generate a rationale arguing for or against the entity as an element of the concept's extension, followed by a final 'positive' or 'negative' answer.

A \textit{large language model} (LLM) is a probabilistic model trained on a natural language corpus that, given a sequence of tokens from a vocabulary occurring in the corpus, generates a continuation of the input sequence. LLMs exhibit remarkable capabilities for natural language processing and generation \parencite{brown2020language}.

Let \( \mathcal{T} \) be the set of sequences of tokens \( T_i = t_1, t_2, \ldots, t_n \) such that \( t_i \) is a token in a predefined vocabulary \( V \). Given a \textit{corpus} \( \mathcal{C} \subseteq \mathcal{T} \), a \textit{language model} \( \mathcal{L_C} \) is a probabilistic model trained on a sample of \( \mathcal{C} \) that defines a distribution over sequences of tokens.  
\begin{equation}
\mathcal{L_C}(T_i) = p(t_1, t_2, \ldots, t_n)         
\end{equation}
is an estimate of the probability of a sequence \( T_i \), given a corpus \( \mathcal{C} \).
A \textit{prompt template} \( P = (T, F) \) is a pair of a sequence of tokens \( T \) and an set of \textit{free} tokens \( F \subseteq \{ f_1, f_2, \ldots, f_n \} \). A \textit{substitution} \( \theta \) with respect to a prompt \( P \) is a set of pairs \( ( f_i, T_i ) \) such that \( f_i \in F \) and \( T_i \in \mathcal{T} \). A \textit{prompt} is a sequence of tokens \( P' \in \mathcal{T} \) such that \( \forall (f_i, T_i) \in \theta \) every occurrence of \( f_i \) in a prompt template \( P \) is replaced with \( T_i \). Given a prompt \( P \), the goal of a language model \( \mathcal{L_C} \) is to generate a sequence of tokens that maximizes the conditional probability under \( \mathcal{L_C} \).
\begin{equation}
T_{\text{out}} = \arg \max_{T} \mathcal{L_C}(T | P)      
\end{equation}
is the output sequence generated by the language model, conditioned on \( P \).

We define a function \texttt{instantiate} such that:
\begin{equation}
    \begin{aligned}
        P' = \texttt{instantiate}(P, \theta)
    \end{aligned}
\end{equation}
where $P$ is a prompt template, $\theta$ is a substitution, and $P'$ is the prompt produced by applying $\theta$ to $P$. Given an language model $\mathcal{L_C}$, we define a function \texttt{classify} as follows:
\begin{equation}
    \begin{aligned}
( T_R, T_\mathbb{B} ) = \texttt{classify}(c, e)      
    \end{aligned}
\end{equation}
where $T_{label(c)}$ is the name of $c$,  $T_c$ is a natural language definition of $c$,  $T_{label(e)}$ is the name of $e$,  $T_e$ is a natural language description of $e$,  $T_R$ is a sequence of tokens that represents a rationale for a classification decision, and $T_\mathbb{B} \in \{ \texttt{positive}, \texttt{negative} \}$ are tokens that represent classification decisions, i.e., whether or not $e$ is in the extension of $c$.

We compute $T_R$ and $T_\mathbb{B}$ as follows:
\begin{equation}
    \begin{aligned}
T_R = \arg \max_{T} \mathcal{L_C}(T | \texttt{instantiate}(P_{rationale\_generation},\theta_0))        
    \end{aligned}
\end{equation}
\begin{equation}
    \begin{aligned}
T_\mathbb{B} = \arg \max_{T} \mathcal{L_C}(T | \texttt{instantiate}(P_{answer\_generation},\theta_1))        
    \end{aligned}
\end{equation}
\begin{equation}
\begin{aligned}
\theta_0 = \{ & (\texttt{\{label\}}, T_{label(c)}),(\texttt{\{definition\}}, T_{c}), \\
& (\texttt{\{entity\}}, T_{label(e)}),(\texttt{\{description\}}, T_{e}) \} 
\end{aligned}
\end{equation}
\begin{equation}
    \begin{aligned}
\theta_1 = \theta_0 \cup \{(\texttt{\{rationale\}}, T_{R})\}        
    \end{aligned}
\end{equation}
given two prompt templates $P_{rationale\_generation}$ and $P_{answer\_generation}$. Table \ref{tab:prompt_templates} displays the specific prompt templates used in our experiments.

\begin{table*}
    \small
    \centering
    \begin{tabular}{p{3cm}|p{8cm}}
        \textbf{prompt template} & \textbf{definition}  \\ 
        \hline
         $P_{rationale\_generation}$ & \texttt{Concept: \{concept\} \newline 
Definition: \{definition\} \newline
Entity: \{entity\} \newline 
Description: \{description\} \newline
   \newline
Using the above definition, and only the information in the above definition, 
provide an argument for the assertion that \{entity\} is a(n) \{concept\}. \newline
     \newline
Rationale:} \\
         \hline
         $P_{answer\_generation}$ &  \texttt{Concept: \{concept\} \newline
Definition: \{definition\} \newline
Entity: \{entity\} \newline
Description: \{description\} \newline
Rationale: \{rationale\} \newline
\newline
Now using the argument provided in the above rationale, answer the question: is \{entity\} a(n) \{concept\}? \newline
Answer 'positive' or 'negative', and only 'positive' or 'negative'.  Use lower case. If there is not enough information to be sure of an answer, answer 'negative'. \newline
  \newline
Answer:}\\
    \end{tabular}
    \vspace*{2mm}
    \caption{Prompt templates used to generate classification procedures.}
    \label{tab:prompt_templates}
\end{table*}

\section{Evaluating classification procedures using knowledge graphs}
\label{sec:4}
Now that we have defined an approach to implementing classification procedures, we turn to the question of how such procedures can be evaluated. To this end, we leverage knowledge graphs as a source of entities to use to evaluate classification procedures for a given concept.

\begin{figure*}
    \includegraphics[width=\columnwidth]{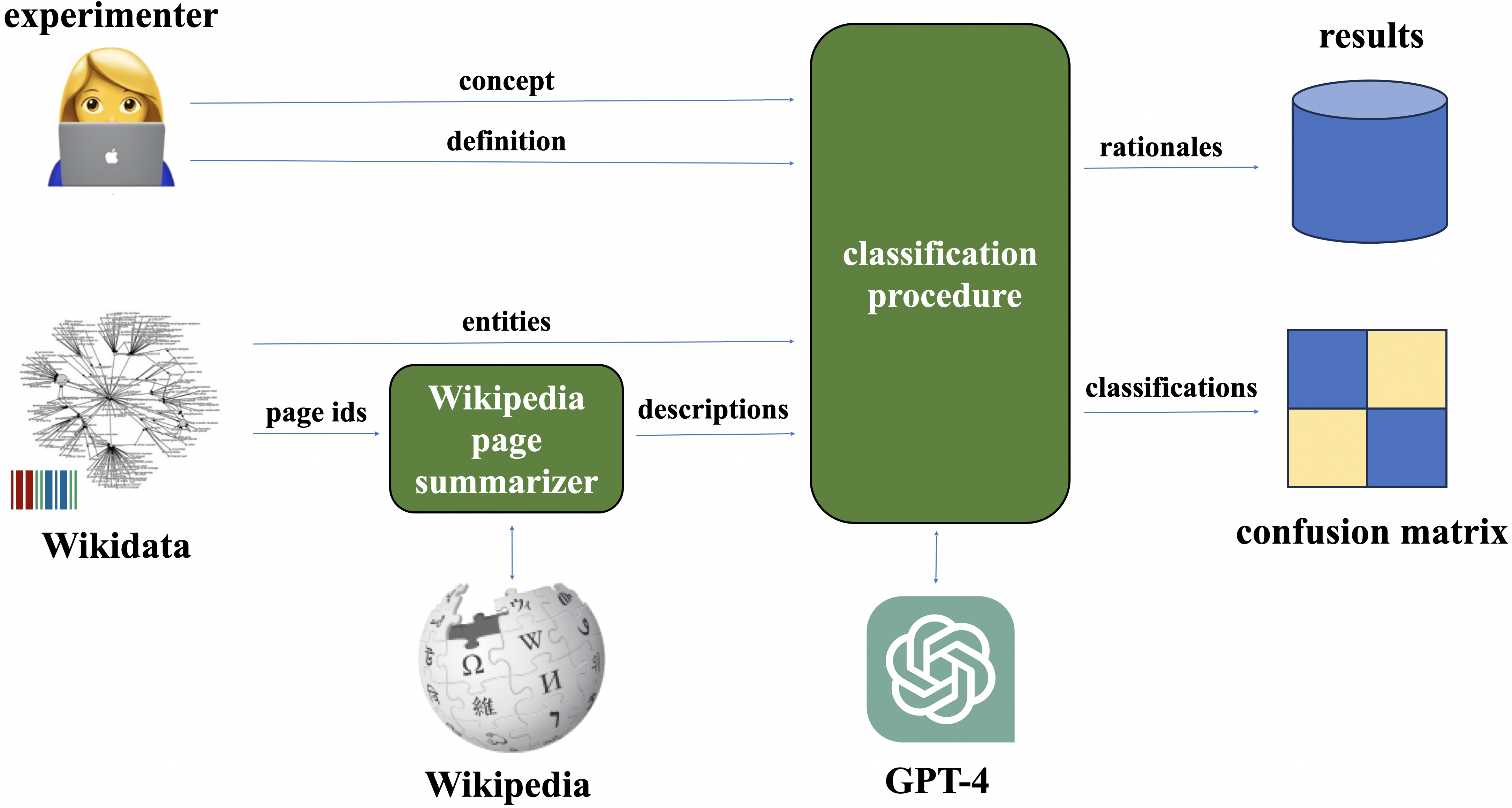}
    \caption{A workflow for evaluating classification procedures using a knowledge graph.}
    \label{fig:evaluation_procedure}
\end{figure*}

A \textit{knowledge graph} represents knowledge using nodes for entities and edges for relations \parencite{10.1145/3447772}. Knowledge graphs are key information infrastructure for many Web applications \parencite{heist2020knowledge}. Following \textcite{angles2020mapping}, we use the RDF data model to describe knowledge graphs. 

Let \( I \) be an infinite set of IRIs (Internationalized Resource Identifiers \parencite{durst2005internationalized}), \( B \) be an infinite set of blank nodes \parencite{hogan2014everything}, and \( L \) an infinite set of literals \parencite{beek2018literally}. 
A \textit{knowledge graph} \( G \) is a set of \textit{triples} \( \{(s, p, o) \mid s \in S, p \in P, o \in O \} \), where \( S \subset I \cup B \) is the set of \textit{subjects} in \( G \), \( P \subset I \) is the set of \textit{properties} in \( G \), and \( O \subset I \cup B \cup L \) is the set of \textit{objects} in \( G \). Let \( \texttt{instanceOf}, \texttt{subClassOf}, \texttt{label} \in P \) denote an instance-of relation, a subclass-of relation, and a label property in \( G \), respectively.
A \textit{concept} \( c \in I \cup B \) is an entity such that \( \exists (s, \texttt{subClassOf}, o) \in  G \mid  s = c \lor o = c \).

We define a function $\texttt{ext}_G(c)$ that computes the \textit{extension in} $G$ of a concept $c\in G$ recursively, such that:
\begin{equation}
\texttt{ext}_G(c) = \bigcup_{i \in \mathop \mathbb{N}} \texttt{ext}_i(c)        
\end{equation}
where
\begin{equation}
    \begin{aligned}
\texttt{ext}_0(c) = \{ e \mid \exists (e, \texttt{instanceOf}, c) \in G \}        
    \end{aligned}
\end{equation}
\begin{equation}
\begin{aligned}
\texttt{ext}_{i+1}(c) = \texttt{ext}_i(c) \cup \{ e \mid e \in \texttt{ext}(c') \land  \exists (c', \texttt{subClassOf}, c) \in G \}
\end{aligned}
\end{equation}

Our evaluation workflow is implemented as follows. We sample positive and negative examples of a concept from a given knowledge graph, using the extension of the concept computed as above as the source of positive examples, and the set difference of that extension and that of a concept related to it by a \texttt{subClassOf} relation as the source of negative examples. We then apply the classification procedure for a given definition of the concept to each example, and compute a confusion matrix from the classifications, which provides performance metrics for the classification procedure. Figure \ref{fig:evaluation_procedure} shows the evaluation workflow, and Algorithm \ref{alg:experiment} describes the procedure in pseudo-code.\footnote{In his 1955 essay "Meaning and synonymy in natural languages" \parencite{carnap1955meaning}, Rudolf Carnap presents a thought experiment wherein an investigator provides a hypothetical robot with a definition of a concept together with a description of an individual, and then asks the robot if the individual is in the extension of the concept. Our evaluation workflow can be viewed as an instantiation of Carnap's experimental framework, with a classification procedure playing the role of Carnap's robot.}

\begin{algorithm}[ht]
    \SetKwFunction{Classify}{classify}
    \SetKwFunction{Ext}{$ext_G$}
    \SetKwInOut{Output}{output }
    \SetKwInOut{Input}{input}
    \Input{a pair of classes $c, d$ from $G \mid (c, \texttt{subClassOf}, d) \in G$}
    \Output{a confusion matrix $M$}
    \BlankLine
    ($TP$, $FP$, $TN$, $FN$) $\leftarrow$ (0, 0, 0, 0)\;\
    $E^+$ $\leftarrow$ a uniform random sample from $\texttt{ext}_G(c)$\;\
    $E^-$ $\leftarrow$ \text{a uniform random sample from} $\texttt{ext}_G(d)$ $\setminus$ $\texttt{ext}_G(c)$\;
    \ForEach{$e \in E^+$}{
        $( T_R, T_\mathbb{B} )$ $\leftarrow$ \Classify{$c,e$}\;\
        \lIf{$T_\mathbb{B} = \texttt{positive}$} {$TP \leftarrow TP + 1$}\
        \lElse{$FP \leftarrow FP + 1$}\
    }
    \ForEach{$e \in E^-$}{
        $( T_R, T_\mathbb{B} )$ $\leftarrow$ \Classify{$c,e$}\;\
        \lIf{$T_\mathbb{B} = \texttt{negative}$} {$TN \leftarrow TN + 1$}\
        \lElse{$FN \leftarrow FN + 1$}\
    }
    $M \leftarrow [ [ TP, FP ], [ FN, TN ] ]$\;
    \BlankLine
    \caption{Evaluation procedure}
    \label{alg:experiment}
\end{algorithm}

\section{Experiments}
\label{sec:5}

Much of what has been written on the theory and practice of conceptual engineering makes reference to two specific paradigmatic projects: the International Astronomical Union's redefinition of PLANET \parencite{iauplanetdraft2006}, and Sally Haslanger's ameliorative analysis of WOMAN \parencite{haslanger2000gender}. We now describe a set of experiments applying the above defined implementation of classification procedures and evaluation workflow to different stipulative definitions of these two concepts.

\subsection{Data}
For our experiments, we evaluated three definitions for PLANET: one from the Oxford English Dictionary (OED) \parencite{oedplanet} and two from the 2006 International Astronomical Union (IAU) General Assembly \parencite{iauplanetdraft2006,assembly2006result}). We evaluated three definitions for WOMAN: one from the OED \parencite{oedwoman}, the definition provided in Haslanger’s 2000 paper \parencite{haslanger2000gender}, and one from the Homosaurus vocabulary of LGBTQ+ terms \parencite{homosauruswomen,cifor2022mediating}. The definitions are shown in Table \ref{tab:definitions}. 

We used the Wikidata collaborative knowledge graph \parencite{vrandevcic2014wikidata} as a source of entities. For PLANET, we sampled 50 positive examples that are instances (P31) of planet (Q634), and 50 negative examples that are instances of substellar object (Q3132741), but not of planet. For WOMAN, we sampled 50 positive examples whose sex or gender (P21) is either female (Q6581072) or trans woman (Q1052281), and 50 negative examples whose sex or gender is either male (Q6581097), non-binary (Q48270), or trans man (Q2449503). For entity descriptions, we use a summary retrieved from Wikipedia of the page corresponding to the Wikidata entity.

We used GPT-4 \parencite{openai2023gpt4} with a temperature setting of 0.1 as the LLM in these experiments. LLM inference API calls were made between 20th and 21st October 2023.

\begin{table*}
    \small
    \centering
    \begin{tabular}{l|l|p{5.5cm}}
        \textbf{concept} & \textbf{source of definition} & \textbf{definition} \\
        \hline
        \textbf{PLANET} & OED \parencite{oedplanet} & Any of various rocky or gaseous bodies that revolve in approximately elliptical orbits around the sun and are visible by its reflected light; esp. each of the planets Mercury, Venus, Earth, Mars, Jupiter, Saturn, Uranus, Neptune, and (until 2006) Pluto (in order of increasing distance from the sun); a similar body revolving around another star. Also: any of various smaller bodies that revolve around these (cf. satellite $n.$ 2a). \\
        \cline{2-3}
        & IAU 2006-08-16 \parencite{iauplanetdraft2006} & A planet is a celestial body that (a) has sufficient mass for its self-gravity to overcome rigid body forces so that it assumes a hydrostatic equilibrium (nearly round) shape, and (b) is in orbit around a star, and is neither a star nor a satellite of a planet. \\
        \cline{2-3}
        & IAU 2006-08-24 \parencite{assembly2006result} & A planet [1] is a celestial body that (a) is in orbit around the Sun, (b) has sufficient mass for its self-gravity to overcome rigid body forces so that it assumes a hydrostatic equilibrium (nearly round) shape, and (c) has cleared the neighbourhood around its orbit. \\
        \hline
        \textbf{WOMAN} & OED \parencite{oedwoman} & An adult female human being. The counterpart of man (see man, $n.^1$ II.4.) \\
        \cline{2-3}
        & Haslanger \parencite{haslanger2000gender} & S is a woman iff (i) S is regularly and for the most part observed or imagined to have certain bodily features presumed to be evidence of a female’s biological role in reproduction; (ii) that S has these features marks S within the dominant ideology of S’s society as someone who ought to occupy certain kinds of social position that are in fact subordinate (and so motivates and justifies S’s occupying such a position); and (iii) the fact that S satisfies (I) and (ii) plays a role in S’s systematic subordination, that is, along some dimension, S’s social position is oppressive, and S’s satisfying (i) and (ii) plays a role in that dimension of subordination \\
        \cline{2-3}
        & Homosaurus \parencite{homosauruswomen} & Adults who self-identify as women and understand their gender in terms of Western conceptions of womanness, femaleness, and/or femininity. The term has typically been defined as adult female humans, though not all women identify with the term 'female' depending on the context in which it is used. \\
    \end{tabular}
    \vspace*{2mm}
    \caption{Definitions for concepts used in the experiments.}
    \label{tab:definitions}
\end{table*}

\subsection{Results}

\begin{table*}
    \small
    \centering
    \begin{tabular}{p{2cm}|p{2.4cm}|p{2.4cm}|p{1.6cm}|p{1cm}|p{1cm}}
        \textbf{concept} & \textbf{definition} & \textbf{Cohen's kappa} & \textbf{F1 macro} & \textbf{FN} & \textbf{FP} \\ 
        \hline
         \textbf{PLANET} & IAU 2006-08-24 & \textbf{0.96} & \textbf{0.98} & 1 & 1 \\
         & IAU 2006-08-16 & 0.94 & 0.97 & 1 & 2 \\
         & OED & 0.92 & 0.96 & 1 & 3 \\
        \hline
         \textbf{WOMAN} & Homosaurus & \textbf{0.96} & \textbf{0.98} & 0 & 2 \\
         & Haslanger & 0.94 & 0.97 & 2 & 1 \\
         & OED & 0.92 & 0.96 & 2 & 2 \\
    \end{tabular}
    \vspace*{2mm}
    \caption{Performance metrics for classification procedures over samples of Wikidata entities (for each concept, N=100, positives=50, negatives=50).}
    \label{tab:summary_performance_metrics}
\end{table*}

Table \ref{tab:summary_performance_metrics} provides a summary of the performance metrics from the experiments. For PLANET, all three classification procedures performed well, with the final (24 August 2006) IAU definition performing best. All three definitions resulted in a classification procedure exhibiting almost perfect agreement with the knowledge graph, as estimated by F1 Macro and Cohen's kappa metrics. For WOMAN, all three classification procedures also performed well, again with high F1 scores, and Cohen's kappa values indicating almost perfect agreement with the knowledge graph. 

\begin{table}[ht]
\centering
\begin{tabular}{l|l|l|l|l|l|l}
\textbf{concept} & \textbf{definition} & \textbf{entity} & \textbf{error} & \textbf{cause} & \textbf{unfaithful} & \textbf{hallucination} \\
\hline
\textbf{PLANET} & OED & 2MASS J03552337+1133437 & FN & KG & no & no \\
& & (613100) 2005 TN74 & FP & KG & no & no \\
& & 2010 GB174 & FP & KG & no & no \\
& & (35671) 1998 SN165 & FP & KG & no & no \\
& IAU 2006-08-16 & 2MASS J03552337+1133437 & FN & KG & no & no \\
& & 2010 GB174 & FP & LLM & no & no \\
& & (35671) 1998 SN165 & FP & KG & no & no \\
& IAU 2006-08-24 & 2MASS J03552337+1133437 & FN & KG & no & no \\
& & 2010 GB174 & FP & LLM & yes & no \\
\hline
\textbf{WOMAN} & OED & Nemesis & FN & LLM & yes & yes \\
& & Brianna Ghey & FN & LLM & yes & yes \\
& & Michelle Rojas & FP & KG & no & no \\
& & Linden A. Lewis & FP & KG & no & no \\
& Haslanger & Waltraud Klasnic & FN & LLM & no & yes \\
& & Michaela Kirchgasser & FN & LLM & no & no \\
& & Michelle Rojas & FP & KG & no & no \\
& Homosaurus & Michelle Rojas & FP & KG & no & no \\
& & Linden A. Lewis & FP & KG & no & no \\
\end{tabular}
\caption{Error analysis. The \textbf{error} column indicates the type of error (FN = false negative, FP = false positive), the \textbf{cause} column indicates the author's opinion as to the source of the error (KG = knowledge graph, LLM = large language model), the \textbf{unfaithful} column is the author's opinion as to whether the classification is unfaithful to the rational, and the \textbf{hallucination} column  is the author's opinion as to whether the rational exhibits hallucination.}
\label{tab:error_analysis}
\end{table}

Table \ref{tab:error_analysis} provides details on the errors made by the classification procedures. We reviewed the errors to determine if a given error arises from the concept's definition or the entity's description. In addition, we reviewed the rationales generated by the classification procedures to determine if their classifications were unfaithful to their rationales, and if they exhibited hallucination, i.e., exhibited incorrect reasoning or false assertions \parencite{ji2023survey}. 

For PLANET, the majority of errors were false positives relating to trans-Neptunian objects, the problematic classification of which was a motivation for the IAU redefinition of PLANET. All of the PLANET classification procedures had 2MASS J03552337+1133437 (Q222246) as a false negative, which was rejected due to its identification as a brown dwarf. Table \ref{tab:IAU_2006-08-24_fp} shows the false positive error for this entity by the classification procedure for PLANET based on the IAU 2006-08-24 definition. This is arguably a case where the knowledge graph is mistaken, for reasons that are described in the classification procedure's rationale. The rationale raises two issues with the knowledge graph's classification. It first correctly asserts that a brown dwarf is not a planet and then, applying a literal interpretation of the IAU definition's problematic use of the term "Sun" instead of e.g. "star" \parencite{sarma2008iau}, notes that it is not orbiting the Sun. It is worth noting that in several other rationales produced by the classification procedure
\begin{table*}
    \footnotesize
    \centering
    \begin{tabular}{p{2.5cm}|p{4.5cm}|p{4.5cm}}
    \textbf{entity} & \textbf{description} & \textbf{rationale} \\
    \hline
    2MASS J03552337+1133437 & 2MASS J03552337+1133437 (2MASS J0355+11) is a nearby brown dwarf of spectral type L5$\gamma$, located in constellation Taurus at approximately 29.8 light-years from Earth. & The given definition cannot support the assertion that 2MASS J03552337+1133437 is a planet. The description of 2MASS J03552337+1133437 identifies it as a brown dwarf, not a planet. Furthermore, the definition requires a planet to be in orbit around the Sun, and there is no information provided that 2MASS J03552337+1133437 orbits the Sun. Therefore, based on the provided definition and information, we cannot argue that 2MASS J03552337+1133437 is a planet. \\
    \end{tabular}
    \vspace*{2mm}
    \caption{A false positive (FP) error for the IAU 2006-08-24 definition of PLANET.}
    \label{tab:IAU_2006-08-24_fp}
\end{table*}

For WOMAN, five of the nine errors were false positives that in the author's judgement reflected errors in the knowledge graph around two specific entities. Table \ref{tab:homosaurus_fns} shows two of these false positive errors occurring in the evaluation of the Homosaurus definition of WOMAN. In these instances, the classification procedure's rationale indicate that the pronouns used and social roles mentioned in the entity descriptions provide evidence that the entities are in fact women.
\begin{table*}
    \footnotesize
    \centering
    \begin{tabular}{p{2.5cm}|p{4.5cm}|p{4.5cm}}
    \textbf{entity} & \textbf{description} & \textbf{rationale} \\
    \hline
    Michelle Rojas & Michelle Rojas (née Pennington; born December 16, 1987) is an American voice actress from Dallas, Texas. Affiliated with Funimation and Bang Zoom! Entertainment, some of her major roles are Tohka Yatogami in Date A Live, Zuikaku in Azur Lane, Kanade Sakurada in Castle Town Dandelion, Mikoto Sakuragawa in Gonna be the Twin-Tail!!, Kō Yagami in New Game!, Nashiro Yasuhisa in Tokyo Ghoul, Maya Sato in Classroom of the Elite, Toka Yada in Assassination Classroom, Shion in That Time I Got Reincarnated as a Slime, Roxy Migurdia in Mushoku Tensei: Jobless Reincarnation, and Yamato in One Piece. & The description of Michelle Rojas does not explicitly state her gender identity. However, it does use female pronouns ("her") to refer to her, which typically aligns with a self-identification as a woman. Furthermore, she has taken on roles that are typically associated with women, such as female characters in various shows. Therefore, based on the information provided and the definition of a woman, it can be argued that Michelle Rojas is a woman. \\
    \hline
    Linden A. Lewis & Linden A. Lewis is a science fiction author, best known for their debut novel The First Sister. Lewis is queer and uses she/they pronouns. & The definition of a woman includes adults who self-identify as women. Linden A. Lewis uses she/they pronouns, which indicates that they may identify as a woman. Therefore, based on the provided definition, it can be argued that Linden A. Lewis is a woman. \\
    \end{tabular}
    \vspace*{2mm}
    \caption{False positive (FP) errors for the Homosaurus definition of WOMAN.}
    \label{tab:homosaurus_fns}
\end{table*}
In contrast, Table \ref{tab:haslanger_fn} shows a false negative error by the classification procedure using the Haslanger definition of WOMAN. This is an error due to the LLM, and the rationale provided exhibits hallucination in that it provides a misinterpretation of clause (ii) of Haslanger's definition. In addition, the classification procedure does not acknowledge linguistic cues in the entity description that the entity is a woman, i.e., the use of the terms \textit{née} and \textit{Landeshauptfrau}.
\begin{table*}
    \footnotesize
    \centering
    \begin{tabular}{p{2.5cm}|p{4.5cm}|p{4.5cm}}
    \textbf{entity} & \textbf{description} & \textbf{rationale} \\
    \hline
    Waltraud Klasnic & Waltraud Klasnic (née Tschiltsch, born 27 October 1945) is an Austrian politician who was Landeshauptfrau (governor) of Styria from 1996 until 2005. & The definition of a woman provided includes three main criteria: (i) having certain bodily features associated with the female biological role in reproduction, (ii) being marked within the dominant ideology of one's society as someone who should occupy certain subordinate social positions, and (iii) the fact that satisfying (i) and (ii) contributes to one's systematic subordination. In the case of Waltraud Klasnic, while the description does not explicitly state her biological features or her societal role, it does mention that she is an Austrian politician who served as the governor of Styria. This implies that she is recognized and accepted in a significant societal role, which suggests that she likely satisfies criteria (ii). However, the description does not provide enough information to definitively assert that Waltraud Klasnic satisfies all the criteria in the definition of a woman. For instance, it does not provide information about her biological features or whether she experiences systematic subordination due to her gender. Therefore, based solely on the provided definition and description, we cannot conclusively argue that Waltraud Klasnic is a woman. \\
    \end{tabular}
    \vspace*{2mm}
    \caption{A false negative (FN) error for the Haslanger definition of WOMAN.}
    \label{tab:haslanger_fn}
\end{table*}

In summary, our method performed well in performing classifications using the provided definitions, as evaluated against the sampled data using F1 Macro and Cohen's kappa as measures of accuracy and alignment. Although our experimental setup assumed the knowledge graph as ground truth, our error analysis shows that, in twelve of the eighteen errors encountered in the experiments, rationales produced by the classification procedures provided arguments with which the author was in agreement that the knowledge graph was itself incorrect, as opposed to the LLM hallucinating or being mistaken in its classification.

\section{Discussion}
\label{sec:6}
We now discuss the above approach and experimental results, raising and addressing a number of potential objections to the use of LLMs for implementing classification procedures. In doing so, we touch on three aspects of theory and practice of conceptual engineering: the definition of its targets, empirical methods, and their practical roles.

\subsection{Classifiers as intensions}
Our work provides evidence that the program suggested by Nado in her Practical Role Account \parencite{nado2023taking} is realizable in practice, in a way that allows conceptual engineers to use stipulative definitions \textit{verbatim} to construct classification procedures. Classification procedures thus realized are "inferentialist devices" \parencite{jorem2024inferentialist}, concrete computational artifacts that can be applied in the context of classification and categorization tasks.

However, in relating classification procedures to Cappelen's proposal of intensions and extensions as targets of conceptual engineering, Nado makes the following distinction:
\begin{quote}
    If a classification procedure is sufficiently consistent and thorough, it will determinately ‘pick out’ a function from worlds to sets of entities within that world. This ‘corresponding function’ will characterize the results of applying the procedure (at the actual world) to each possible world. The output of a procedure’s corresponding function when we input a given world is the set of members, at that world, of the category that the classification procedure generates. \ldots
    Some such procedures – ‘well-defined’ ones – will determinately pick out an intension-like function from worlds to sets of entities, and multiple procedures may pick out the same function. Non-well-defined procedures will generate either incomplete or inconsistent classifications, and thus will not determinately fix a world-to-set function. Nonetheless, some non-well-defined procedures may be perfectly reasonable tools for classification. \parencite[13]{nado2023taking}
\end{quote}

Because our definition of \texttt{classify} does not provide a way to use a description of a possible world to provide additional context in generating a classification decision, classification procedures as we have implemented them are, by Nado's account, non-well-defined. We assert that our experiments provide evidence that our approach shows that, in spite of this, classification procedures defined using our method are "perfectly reasonable tools".

That said, there is a way to make our classification procedures well-defined in the above sense. Consider an intensional semantics \parencite{von2011intensional} for a first-order language, where $W$ and $D$ are non-empty sets of possible worlds and individuals, respectively. If we extend the definition of \texttt{classify} to take a natural language description of a possible world as an additional argument, then we can define an \textit{intension} $\llbracket c \rrbracket$ of a concept $c$ as follows: for each $w \in W$ and $e \in D$, $e \in \llbracket c \rrbracket(w)$ if and only if $( T_R, T_\mathbb{B} ) = \texttt{classify}(c, e, w)$ and $T_\mathbb{B} = \texttt{positive}$. This extension of our method is related to similar proposals for defining intensions as classifiers; e.g., \textcite{muskens2005sense} defines intensions as logic programs, and \textcite{larsson2015formal} defines intensions using perceptron-based classifiers.

\subsection{Trustworthiness}
We have seen in our experimental results that the rationales produced by our classification procedures in some instances exhibit hallucinations. Therefore an objection could be made to our approach based on this observed behavior. 

A large amount of work has been performed on different prompt engineering approaches to reduce hallucination in general to improve the ability of LLMs to generate natural language that exhibits consistent and sound reasoning \parencite{wei2022chain,marasovic2021few,creswell2022selection,madaan2023self,yao2023tree,miao2023selfcheck,besta2023graph,dhuliawala2023chain}. Additionally, a variety of approaches to hallucination detection as a means of flagging when an LLM is producing them have been put forward \parencite{ji2023survey,huang2023survey,allen2024shroom}. 
Additional work specifically addresses the reliability and faithfulness of rationales \parencite{ye2022unreliability}, as well as evolving approaches to rationale refinement, exploration and verification \parencite{huang2022towards}. An additional concern stems from evidence that that humans can be misled by erroneous rationales generated by LLMs \parencite{si2023large,heersminkphenomenology}. A number of researchers have proposed that the challenges in this research area are such that the concept of interpretability of LLMs and machine learning models in general needs to be reconsidered \parencite{singh2024rethinking,jacovi2020towards}. 

 Research into the mitigation of hallucination is at an early stage. The current continued rapid growth in LLM capabilities makes the trustworthiness of LLMs a moving target. We are optimistic that conceptual engineers, working with an modicum of epistemic vigilance \parencite{sperber2010epistemic}, can fruitfully apply LLM-based classification procedures in conceptual engineering projects in a manner touched on in Section \ref{sec:empmthds}, even given these concerns\footnote{After all, "Philosophers are (usually) competent natural language speakers and especially keen to subtle differences in meaning." \parencite[172]{justus2012carnap}}.

\subsection{Groundedness}
Another objection arises if one maintains that an understanding of the meaning of the word or phrase used to communicate a concept is important for effective conceptual engineering, as it is an open question at this time as to whether or not LLMs capture and use meaning \parencite{bender2021dangers,lederman2024language}. 

\textcite{mandelkern2024language} argue that LLMs are indirectly verbally grounded in the language present in their training corpora, and thus capable of a limited form of meaning. Beyond that, it is also the case that our method can be said to ground the LLM through the prompt, by incorporating language provided by the conceptual engineer in the definition of the concept, and by the knowledge graph in the summary description of the entity presented during evaluation. This is the approach used in retrieval-augmented generation \parencite{gao2023retrieval} and knowledge-graph-enhanced LLMs \parencite{dai2024counter} to reduce hallucination and improve accuracy.

The question of the groundedness of LLMs is a fascinating one, but from the perspective of Nado's Practical Role Account, it is not clear that this question has any bearing on the utility of our approach:
\begin{quote}
    Though there is a fairly strong correlation between words and procedures, conceptual engineering isn’t about what our words should mean, or even about how we should use our words. It is about \textit{how we should classify}. \ldots If we want our conceptual engineering interventions to affect how people infer and behave, then changing the meaning of a term seems a rather inefficient stratagem. Why not target the classificatory practice directly? \parencite[1993]{nado2023taking}
\end{quote}

Our approach indeed targets the classificatory practice directly, and our experimental results show evidence of useful levels of performance.

\subsection{Empirical methods}
\label{sec:empmthds}
We assert that the evaluation procedure we have defined shows how a conceptual engineering project can incorporate an empirical, data-driven activity \parencite{andow2020fully}. Applying classification procedures to large numbers of positive and negative examples of a concept's extension can help conceptual engineers evaluate different definitions for a concept at a scale that "armchair-based conceptual engineering" \parencite{landesconceptual} cannot. Rationales generated by classification procedures can help conceptual engineers refine their definitions. This raises the possibility that generative AI assistants \parencite{weisz2023toward} could support philosophers in the conduct of conceptual engineering projects.

In addition, recent work on using LLMs as models of human linguistic behavior or judgment, and their use in simulating linguistic subpopulations \parencite{aher2023using,argyle2023out,horton2023large,dillion2023can,simmons2023large},
further suggests that our proposed method could be combined with that work to yield a corpus method for experimental philosophy \parencite{fischer2022projects,sytsma2023ordinary}. 

\subsection{The implementation problem}
\textcite{cappelen2018fixing} and others have argued that conceptual engineering is difficult, as it is hard to see how the natural language (re)definition of a concept can be effectively adopted by a population of human speakers. This has come to be known as the implementation problem \parencite{cappelen2018fixing,jorem2021conceptual}. 
We assert that our approach, used as a means for semantically aligning intensional knowledge expressed in natural language and extensional knowledge represented in a knowledge base \parencite{allen2024evaluating}, can play a practical role in providing a new set of success conditions for conceptual engineering \parencite{andow2021conceptual,pinder2022haslanger}.

Knowledge bases such as Wikidata have an impact on society by virtue of their use in online search, discovery, and recommendation \parencite{peng2023knowledge}. 
Using classification procedures to evaluate and improve the alignment between natural language definitions of concepts and the representation of their extensions in knowledge graphs can be of practical value in knowledge graph refinement, which is the process of improving an existing knowledge graph by adding missing knowledge or identifying and removing errors \parencite{paulheim2017knowledge}. Engineering concepts represented in such resources using the above method can aid understanding within a specific linguistic subgroup, i.e., the users of applications built on top of such knowledge bases, as proposed in \parencite{matsui2024local}. As an example use case closely related to the experiments described above, the Wikidata community is working to improve the modeling of gender in Wikidata \parencite{wikidatagender2023}; we hypothesize that our approach would be useful in efforts of that sort.

Related to the task of knowledge graph refinement are socially responsible data management \parencite{stoyanovich2022responsible} and data governance \parencite{khatri2010designing}. \textcite{khatri2010designing} describe principles for data governance, touching on issues of the alignment of natural language concepts and their realization in databases. These concerns are echoed in the FAIR principles \parencite{wilkinson2016fair}, specifically with respect to the requirement for clear documentation of metadata that aligns natural language concepts and metadata in scientific data resources. More recently, \textcite{vogt2024fair} have proposed additional to the FAIR principles to specifically address the issue of semantic interoperability. Given the increasing use of knowledge graphs in scientific research and commercial applications, these principles are important to apply in the context of knowledge graph creation and refinement. We believe that our approach could be useful in this context as well.

\section{Limitations}
\label{sec:7}
A limitation of our work is its reliance on a specific, proprietary LLM inference API \parencite{openai2023gpt4}, which raises transparency, reproducibility and safety concerns \parencite{bender2021dangers, hu2023prompting}. Reproducing these experiments using other inference APIs, including ones based on open-source or open weight LLMs, would provide useful information with respect to the variation in performance due to the use of other LLMs. Recently, we have shown that our approach, applied to the task of knowledge graph refinement, has good performance across seven different LLMs  \parencite{allen2024evaluating}.

Another limitation in our experiments is that error analysis was performed solely by the author. More reviewers, reviewing a larger set of examples and classifications, would provide a stronger statistical estimate of the level of agreement between human evaluators, the classification procedure, and the knowledge graph.

Finally, we did not investigate the effect of two specific choices made in the prompt engineering of the classifier. First, the $P_{rationale\_generation}$ prompt explicitly provided instruction to ignore background information present in the training corpus for the LLM in considering the intensional definition, and second, the $P_{answer\_generation}$ prompt explicitly provided instruction intended to ensure that a binary classification was made. In the case of the latter, another implementation could instead use a ternary-valued logic, such as a weak Kleene logic \parencite{beall2016off,ciuni2019semantical,zamperlin2019intensional}, with an additional truth value of \texttt{undefined}. Ablation studies would provide insight into the validity of these two prompt design choices.

\section{Conclusion}
\label{sec:8}
In this work, we have shown how to construct a conceptual engineering target as a computational artifact, and apply it to provide an empirical method for use in conceptual engineering projects. We view this as an initial step in an investigation of the potential utility of large language models in the practice of conceptual engineering.

Much has been written of late on the impact that large language models will on society. There is clearly much work to be done to address issues of their trustworthiness, safety, ethics, and environmental impact. That being said, we hope that the work here suggests that LLMs, through their use in the context of ameliorative and normative projects of conceptual engineering \parencite{haslanger2000gender,kohler2024conceptual}, can play a positive role in the future.

\section*{Acknowledgements}
The author wishes to thank Paul Groth, Corey Harper, Filip Ilievski, Jürgen Lipps, and Lise Stork for useful conversations and suggestions with respect to the topics discussed above, and Nathaniel Gan and Nikhil Mahant for their thorough review and valuable feedback, which improved the manuscript.

\printbibliography

\end{document}